\documentclass[letterpaper, 10 pt, conference]{ieeeconf}  

\IEEEoverridecommandlockouts                              

\usepackage{color}
\usepackage{epsfig}
\usepackage{graphicx}
\usepackage{algorithm,algorithmic}

\usepackage{adjustbox}
\usepackage{array}
\usepackage{booktabs}
\usepackage{colortbl}
\usepackage{float,wrapfig}
\usepackage{framed}
\usepackage{hhline}
\usepackage{multirow}
\usepackage{makecell}
\usepackage[percent]{overpic}

\usepackage{amsmath,amsfonts,amssymb}
\usepackage{amsthm} 
\usepackage{bm}
\usepackage{nicefrac}
\usepackage{microtype}
\usepackage{contour}
\usepackage{courier}
\usepackage{siunitx}

\usepackage{changepage}
\usepackage{extramarks}
\usepackage{fancyhdr}
\usepackage{lastpage}
\usepackage{setspace}
\usepackage{soul}
\usepackage{xspace}
\usepackage{cuted}
\usepackage{fancybox}
\usepackage{afterpage}
\usepackage{gensymb}

\usepackage[breaklinks=true,colorlinks,backref=True]{hyperref}
\hypersetup{colorlinks,linkcolor={black},citecolor={MSBlue},urlcolor={magenta}}
\usepackage{url}
\usepackage{quoting}
\usepackage{epigraph}

\usepackage{enumerate}
\usepackage{paralist,tabularx}
\usepackage{comment}
\usepackage{pdfpages}
\usepackage{caption}  

\usepackage{pifont}

\usepackage{MnSymbol}
\usepackage{bbm}

\usepackage[para]{footmisc}
\usepackage{cite}

\usepackage{censor}

\makeatletter
\DeclareRobustCommand\onedot{\futurelet\@let@token\@onedot}
\def\@onedot{\ifx\@let@token.\else.\null\fi\xspace}

\makeatother

\definecolor{MyDarkBlue}{rgb}{0,0.08,1}
\definecolor{MyDarkGreen}{rgb}{0.02,0.6,0.02}
\definecolor{MyDarkRed}{rgb}{0.8,0.02,0.02}
\definecolor{MyDarkOrange}{rgb}{0.40,0.2,0.02}
\definecolor{MyPurple}{RGB}{111,0,255}
\definecolor{MyRed}{rgb}{1.0,0.0,0.0}
\definecolor{MyGold}{rgb}{0.75,0.6,0.12}
\definecolor{MyDarkgray}{rgb}{0.66, 0.66, 0.66}
\definecolor{MyPink}{rgb}{1, 0.75, 0.79}
\definecolor{GreenStarColor}{rgb}{0.54, 0.84, 0.41}
\definecolor{MSBlue}{rgb}{0, 0.35, 0.49}

\def\OURS{ToFTouch\xspace}

\newif\ifblind
\blindtrue  

\newcommand{\redacted}[1]{\ifblind\rule{2cm}{0.4em}\else#1\fi}
\title{\huge \bf \OURS: Exploring Proximity and Tactile Sensing for Robotic Manipulation}

\blindfalse

\author{
\redacted{
Eric T.~Chang$^{1,2,*}$,
Bianca~Aumann$^{1,*}$,
Zhanpeng~He$^{1}$,
Mohamed~Elnoor$^{1}$,
Thomas~Liang$^{1}$,\\
Matei~Ciocarlie$^{2}$,
Nima~Fazeli$^{1}$,
Joshua~Migdal$^{1}$,
Tye~Brady$^{1}$,
Ta\c{s}kin~Pad{\i}r$^{1}$
}
\thanks{
\redacted{
$^{*}$ indicates joint first authorship.
}
}%
\thanks{
\redacted{
\newline $^{1}$Amazon Robotics, MA, USA. \{\texttt{chanerid}, \texttt{biancjur}, \texttt{hzp}, \texttt{melnoor}, \texttt{sliangt}, \texttt{jmigdal}, \texttt{nfz}, \texttt{bradytye}, \texttt{ptaskin}\}@amazon.com
}
}%
\thanks{
\redacted{
\newline$^{2}$Mechanical Engineering Department, Columbia University, NY, USA. 
\{\texttt{eric.chang}, \texttt{matei.ciocarlie}\}@columbia.edu. 
Work done during a co-op at Amazon Robotics.
}
}
}

\begin{document}



\maketitle  

\thispagestyle{empty}
\pagestyle{empty}

\begin{abstract}
Pre-touch sensing, or sensing information about an object before contact, is an attractive modality for robot hands as it can allow a robot to anticipate and respond to a contact before it happens. In this work, we introduce ToFTouch, a multimodal tactile sensor combining three multizone (4$\times$4 resolution) time-of-flight sensors with an optics-based pressure sensing array. We mount this sensor on a parallel jaw gripper and demonstrate its utility on three distinct tasks: (i) pre-grasp adjustment, (ii) item extraction from clutter, and (iii) item insertion into tight spaces. For pre-grasp adjustment, ToF feedback reduced cycle time by 7.4$\%$ by enabling anticipatory gripper closure during approach. For extraction, ToF sensing enabled successful retrieval across all tested depth offsets ($\pm$2 cm) deviating from the nominal pick point, including conditions where baseline approaches failed, with standard deviations below 0.4 cm. For insertion into dense clutter, we explore the proximity and tactile information provided from ToFTouch during an insertion trajectory. The results demonstrate the utility of multizone ToF sensing in a manipulation environment.
\end{abstract}
\section{Introduction}



Humanoid manipulation in complex, unstructured settings ranging from warehouse logistics to the home require a sensor suite that perceives the environment before, during, and after contact.
Traditional tactile sensors are fundamentally reactive: they allow a controller to respond only after contact has occurred.

Proximity, or pre-contact, sensing bridges the gap between seeing an object and touching it, allowing robots to anticipate contact before it happens. For example, pre-contact sensing can allow a robot to pre-shape a gripper to the geometry of an object, avoid undesired contact with neighboring objects or people, and detect the environment surrounding a contact \cite{navarro_proximity_2021}. Furthermore, pre-contact data reduces effective latency \cite{designmultimodalfingertipsensor}, since a control loop can respond to a contact before or as it occurs. While vision systems can serve similar functions, they are subject to occlusion, lighting, latency, and bandwidth limitations.

\begin{figure}[htbp]
\centering
\includegraphics[width=\linewidth]{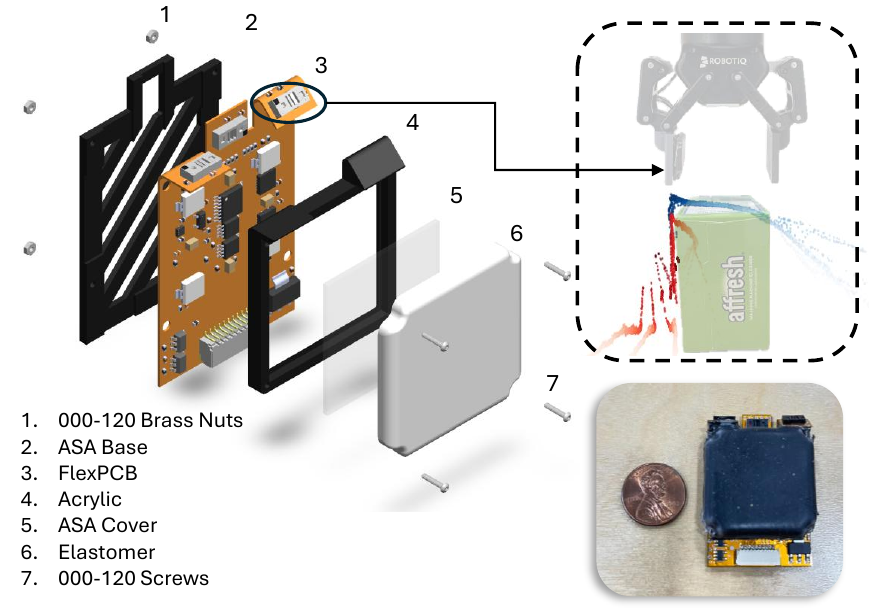}
\caption{Exploded view of ToFTouch. The lower inset compares the sensor size to a U.S. penny. The upper inset shows an example of ToF data accumulated over time during a grasp trajectory with the sensor mounted on a parallel gripper. Red points are from the sensor pointing down and blue points are from the sensor pointing at a 45$\degree$ angle towards the object.}
\label{fig:toftouch}
\end{figure}

Miniature time-of-flight (ToF) sensors offer an attractive implementation to pre-contact sensing. They are small, inexpensive, and available off-the-shelf, providing direct distance measurements with low latency (fewer than eight milliseconds \cite{designmultimodalfingertipsensor}) and low power consumption. These sensors achieve faster sampling rates than many cameras (e.g. 100 Hz). Using infrared emissions, these sensors can operate in varied lighting conditions. Their compact surface mountable form factor, as small as approximately \SI{2}{\milli\meter} $\times$ \SI{4}{\milli\meter}, enables strategic placement in multiple locations across a robot, even in spaces where traditional cameras cannot fit \cite{sifferman_efficient_2025}. This distributed sensing approach allows robots to position ToF sensors where needed, including fingertips, gripper surfaces, and other areas prone to contact, providing comprehensive coverage without the expense, form factor, and occlusion challenges of high-resolution sensors.

Multizone ToF sensors are descendants of single-point ToF sensors that provide denser spatial information, typically at a slight cost to the sampling rate. By dividing the field of view into a grid (e.g., $4\times4$ or $8\times8$), these sensors produce a coarse depth map rather than a single distance value, enabling direct sensing of \emph{local geometry} in the immediate vicinity of contact. Their compact form factor allows integration into robotic fingertips, effectively providing a rich local depth measurement from the fingertip's point of view.
This fingertip point of view is especially valuable for multi-fingered and humanoid hands. The many finger degrees of freedom and the wide variety of reachable hand poses tend to occlude wrist- or palm-mounted cameras, whereas a fingertip-mounted ToF sensor moves with the fingertip and retains a local view of the region around contact. We do not study the multi-fingered hand case in this work, but we expect fingertip point-of-view proximity sensing to become increasingly useful as hand dexterity grows.
Despite these advantages, multizone ToF sensors remain largely unexplored in robotic manipulation research.

We present ToFTouch, a novel tactile sensor that integrates three multizone ToF proximity sensors with an optics-based force sensing array on a 3$\times$3 \SI{}{\centi\meter} patch. The ToF sensors are oriented at different angles to provide comprehensive spatial coverage, each delivering $4\times4$ depth arrays at \SI{60}{\hertz}, while the tactile array operates at 300 Hz for normal force detection. This combination enables robots to anticipate and respond to nearby objects before contact occurs while simultaneously monitoring contact forces. The fusion of these sensing modalities addresses the critical gap between pre-contact awareness and tactile feedback. The sensor's compact design allows integration directly onto gripper fingers, providing local environmental perception even when cameras are obstructed.

Through systematic characterization across three manipulation tasks: (i) pre-grasp adjustment, (ii) item extraction from clutter, and (iii) dense packing insertion, we explore the utility of fingertip ToF sensing in manipulation. Our hardware contribution provides the robotics community with an integrated multizone ToF and tactile sensing platform, along with insights into how proximity information complements traditional sensing modalities for manipulation in cluttered environments.

The primary contributions of this work are as follows.
\begin{itemize}
\item We present ToFTouch, a compact multimodal sensor that, to our knowledge, is the first to integrate \textit{multizone} ToF sensors with a tactile sensing array. The sensor provides 48 proximity zones of 1-\SI{400}{\centi\meter} at \SI{60}{\hertz} and 16-channel tactile feedback at \SI{300}{\hertz}. 
\item We mount the sensor on a robotic fingertip and characterize the sensor across three manipulation tasks: pre-grasp adjustment, extraction from clutter, and insertion for dense-packing. Through these tasks, we demonstrate that fingertip-mounted multizone ToF sensing enables anticipatory grasp control (7.4\% cycle time reduction), depth correction under uncertainty, and real-time environment mapping with point cloud information.
\end{itemize}




\section{Related Work}

Tactile sensing has been crucial for contact-rich manipulation tasks, providing essential feedback about contact forces and object interactions. Traditional applications of tactile sensing focus on characterizing properties of objects during grasping, such as grip stability and object features \cite{designmultimodalfingertipsensor, interfingerManipulation}. However, our work explores a different paradigm where tactile sensing is used in complement with pre-contact sensing in order to anticipate contacts before they happen and detect other objects or surfaces in the environment.

ToF sensing has emerged as a promising proximity modality, offering pre-contact information that can reduce effective control latency \cite{designmultimodalfingertipsensor}. 
Multiple single point ToF sensors placed in different orientations on a fingertip can compose a comprehensive understanding of the environment around the end of arm tool (EoAT) \cite{designmultimodalfingertipsensor}, with some work specifically utilizing forward and downward-facing sensors to detect both target objects and support surfaces \cite{roboticGraspingProximity}. 
Arrays of ToF sensors in robotic fingertips can also enable in-hand pose estimation during manipulation, providing richer spatial information compared to tactile-only sensing \cite{opticalProximityPose}. 
Other work has integrated proximity sensing into the control of fingers on an EoAT \cite{integratedControl} and explored using ToF sensors for grasping, tool manipulation, and surface following tasks \cite{fewShotTransfer}. 
Our work focuses on multizone ToF sensors and the information they can provide during a dense insertion task.

Various approaches have been developed to combine proximity and touch sensing compactly, including capacitive-based bimodal sensors for robotic fingertips \cite{tactileProximity}. Capacitive based proximity sensing tends to be better suited for human interaction due to the dependence of the dielectric constant of the sensed object \cite{navarro_proximity_2021}.
Other approaches include embedded ToF sensors within elastomers to enable both proximity and force sensing from a single sensor \cite{improvedProximity, integratedForce}, camera-based systems that provide both proximity and tactile feedback \cite{fingerSTS, combiningFingerVision}, and vision-enhanced proximity sensors distributed across both fingertips and palm surfaces \cite{lookToTouch}. Sensors that combine optical proximity sensing within the elastomer require careful design to prevent surface features or reflected light from interfering with the proximity measurement. 
We opt for separating the ToF sensors from the elastomer and explore the combination of multizone ToF sensors with optics-based tactile sensing.

Dense packing and precise insertion remain challenging robotic tasks, particularly when visual occlusion limits traditional camera-based approaches. While some approaches employ force control \cite{peginhole}, utilize tactile feedback to predict and correct placement errors \cite{tactileInsertion}, or propose safety to avoid collisions with surrounding objects during insertion \cite{fan}, these strategies can limit performance in tight spaces. The application of proximity sensing to manipulation tasks in pre-contact sensing for picking and insertion tasks, especially in visually occluded environments, remains largely unexplored. Our work introduces multizone ToF sensing combined with tactile feedback for manipulation tasks where visual occlusion is common. 
\section{Hardware Design}
Our system comprises a novel multimodal sensing module, called ToFTouch, that combines proximity and tactile sensing capabilities on a flexPCB contained in a plastic housing with an acrylic and elastomer layer on top (Figure \ref{fig:toftouch}). The module integrates three multizone ($4\times4$ resolution) ToF sensors and a $3\times3$ \SI{}{\centi\meter} optics based normal force sensing patch. The entire sensor is $45\times30\times10$ \SI{}{\milli\meter}.


\subsection{Time-of-Flight Sensing}
The system utilizes three VL53L8CX ToF sensors (STMicorelectronics), each providing a 4x4 depth array at \SI{60}{\hertz} with a 65$\degree$ diagonal field of view. The sensors are rated up to \SI{400}{\centi\meter}, and we find readings to be unreliable at $<$1-\SI{2}{\centi\meter}. The sensors are strategically oriented at 0$\degree$, 45$\degree$, and 90$\degree$ from vertical, providing graduated coverage of the manipulation workspace. This configuration enables broad spatial coverage while maintaining a compact form factor.

\subsection{Tactile Sensing}
The tactile sensing component consists of a flat 3$\times$3 \SI{}{\centi\meter} patch utilizing an optics based tranduction method \cite{piacenza_sensorized_2020}. The sensor contains an array of four LEDs and four photodiodes (PD). These elements are covered by an acrylic sheet and transparent elastomer which acts as a waveguide for light emitted from the LEDs. The transparent elastomer is coated with a reflective surface and opaque surface for reflecting internal light and blocking out external light, respectively. During operation, we sample each LED-PD pair independently. We light up each LED to a pre-calibrated brightness, measure the response from the corresponding photodiode, and repeat for each LED-PD pair. The light response detected by the photodiodes corresponds to deformation imparted on the elastomer surface. The system samples each LED-PD pair at \SI{300}{\hertz}, generating a 16-dimensional vector with careful timing control to ensure signal stability between measurements. Previous work \cite{piacenza_sensorized_2020} shows that explicit force information can be extracted from this sensor paradigm using a learned model, however we examine only the raw signals in this work.

\subsection{Electronics Integration}
The sensing module's electronics are built around a flexible PCB design that accommodates the folded arrangement of ToF sensors. The system operates from a \SI{6}{\volt} input supply, with separate regulated power domains for the ToF sensors and analog sensing components to ensure signal integrity. Level shifters manage the communication between the 3.3V microcontroller logic and 1.8V ToF sensor logic. For precise LED brightness control, the system employs a 16-bit DAC. The four PD signals are conditioned through high-gain transimpedance amplifiers ($R_f=$ \SI{200}{\kilo\ohm}, $C_f=$ \SI{6.2}{\pico\farad}), followed by a high-speed ADC signal acquisition. All ToF sensors and the ADC interface with a Teensy 4.1 microcontroller through SPI over a 10-pin JST connector. The microcontroller communicates with a PC using micro-ROS \cite{belsare2023microrOS}.

\subsection{Fabrication}
The sensor module implements a layered construction approach to integrate the sensing elements while ensuring robust operation. The flexible PCB is reinforced with a stainless steel stiffener to provide structural integrity but the area under the ToFs is not reinforced to allow the flexibility for the sensors to fold into varying angles. An ASA plastic housing, chosen for its thermal stability, encases the electronics and incorporates ventilation features for thermal management. 

To cast the base elastomer layer, we apply two coats of silicone primer to the acrylic, allowing each to dry between layers. We mix and degas Smooth-On Solaris (mixed at a 0.8:1 A:B ratio), then pour into a mold containing the acrylic sheet, allowing to cure for 24 hours at room temperature. After curing and removing from the mold, we adhere the combined acrylic and elastomer unit into the ASA mold with Smooth-On SIL-poxy adhesive. The assembly is dipped twice into the reflective coating (EcoFlex 00-10 with 10\% Cast Magic Silver Bullet powder), allowing to cure until set between layers. Then, we dip the elastomer into an additional two layers of opaque coating (EcoFlex 00-10 mixed with 3$\%$ black SilcPig silicone pigment) to shield the optical sensing elements from ambient light interference. The elastomer cartridge is then assembled to the flexPCB with screws. The ToF sensors are folded onto flat ASA surfaces and adhered with a quick cure epoxy. The complete assembly attaches directly to a robotic gripper surface using double-sided adhesive tape, ensuring secure mounting while maintaining the option for replacement or maintenance.
\section{Sensor Validation}
\begin{figure}[htbp]
\centering
\includegraphics[width=\linewidth]{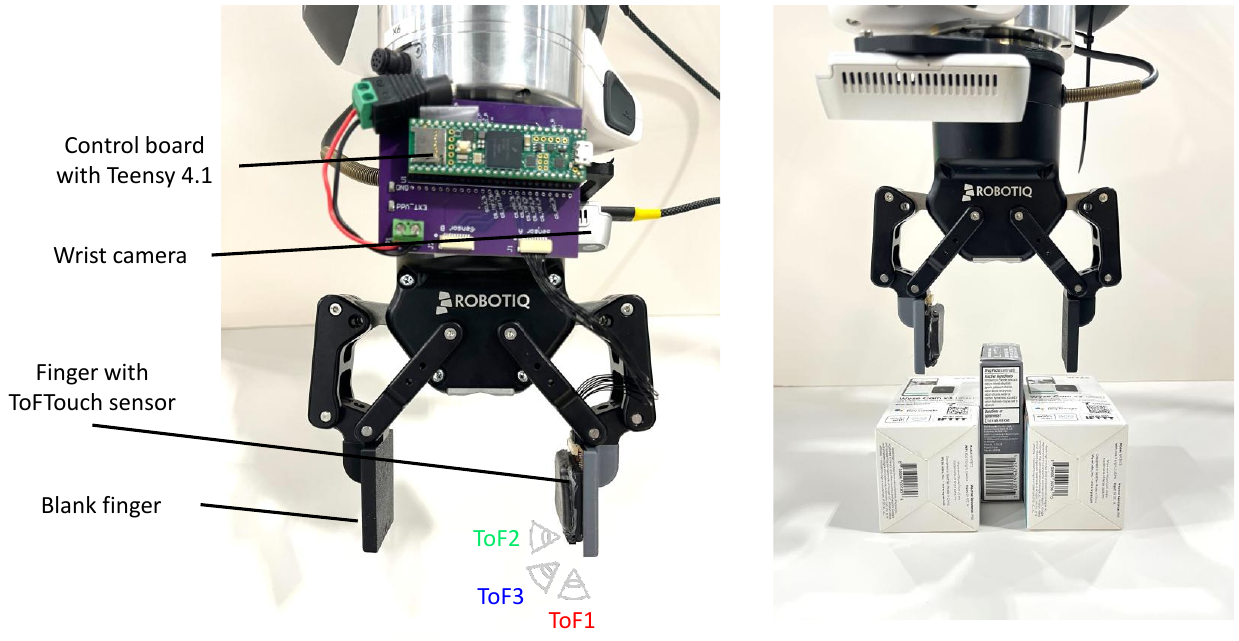}
\caption{Left: ToFTouch and its key elements mounted on one finger of a Robotiq parallel jaw gripper. Right: The parallel jaw gripper above the environment used for extraction experiments.}
\label{fig:robotiq}
\end{figure}

Figure \ref{fig:robotiq} shows the ToFTouch sensor mounted on the inside surface of a Robotiq parallel gripper finger. The ``eye" symbols show the ToF sensor locations and directions. The tactile patch is oriented towards the grasp such that it will measure contact forces with an item in hand. The gripper is mounted on a Franka FR3 arm which is used in all experiments.

\begin{figure}[htbp]
\centering
\includegraphics[width=\linewidth]{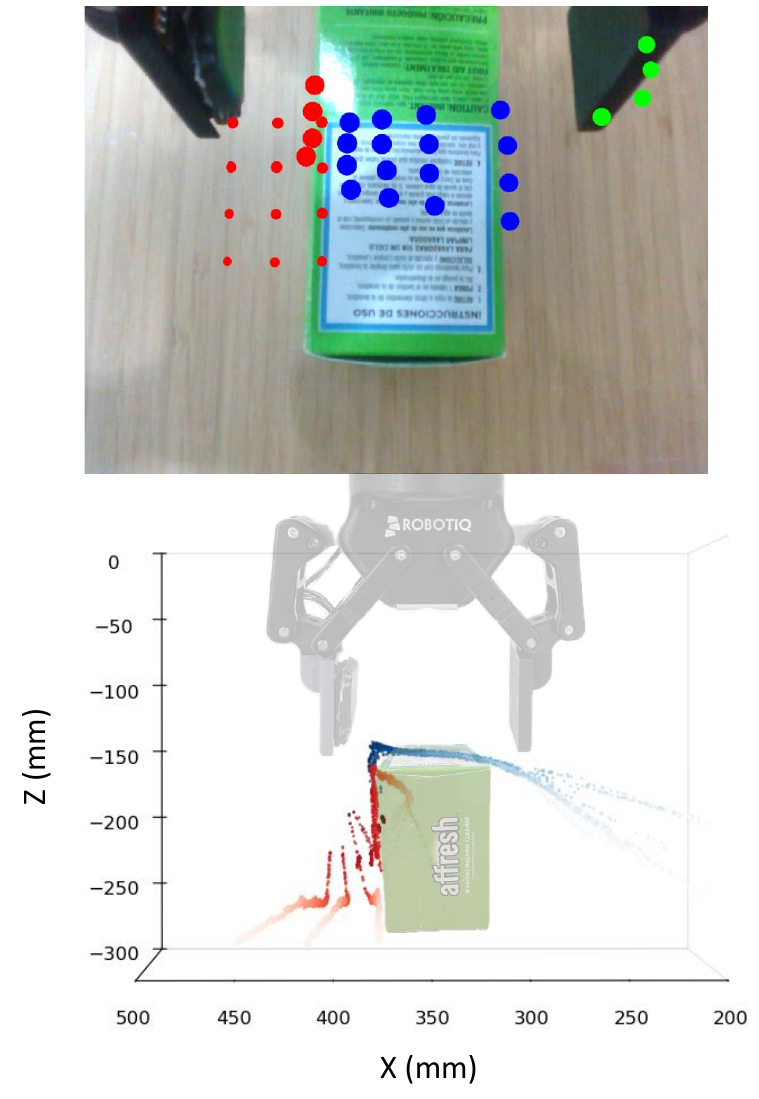}
\caption{Visualization of ToF points at one time step (top) and accumulated over time (bottom). Point color corresponds to data from the ToF sensors as labeled in Figure \ref{fig:robotiq}. Green points from ToF2 are omitted from the bottom image.}
\label{fig:grasp}
\end{figure}

Utilizing this setup, Figure \ref{fig:grasp} illustrates the depth measurements from the ToF sensors at a single timestep, superimposed on the field of view of an on-arm camera for visualization purposes only; the camera is not used for control or feedback in our experiments. The diameter of the data points is scaled by value: larger circles indicate readings closer to the sensor. Red, blue, and green show readings from the three unique multizone ToF sensors. The bottom half of Figure \ref{fig:grasp} shows accumulated measurements over time, creating a localized point cloud of the detected object. The red data from ToF1 identifies the table surface and edge of the box whereas ToF3 shows readings across the top surface of the box.

We designed three experiments to evaluate the utility of ToFTouch:
\begin{enumerate}
    \item \textbf{Pre-grasp adjustment:} Adjust the gripper width while approaching a target item.
    \item \textbf{Extraction:} Pick an item from clutter using only ToF feedback to reach the correct depth.
    \item \textbf{Insertion:} Insert an item into tight clutter using ToF to inform about available space and tactile sensing to monitor extrinsic contacts.
\end{enumerate}

\subsection{Pre-Grasp Adjustment}
In this experiment, we demonstrate that ToF sensing can be used to initiate gripper closure during approach, reducing cycle time compared to waiting until the EoAT reaches the pick point. As humans, we instinctively perform this adjustment with our hands; when reaching for an object, we anticipate the contact geometry and begin closing our hands as we approach objects rather than waiting until contact. This pregrasping strategy can reduce overall grasp cycle time while maintaining reliable object acquisition.

For this experiment, we assume a known pick point and known object geometry. In the baseline condition without ToF sensing, the gripper travels to the pick point at the known depth before initiating closure. In the ToF-enabled condition, the gripper begins closing when the ToF sensor detects the table surface at a threshold distance of \SI{130}{\milli\meter}, allowing closure to occur during the approach motion. For both conditions, we perform ten trials and record the time from movement start until the gripper finishes closing and the robot reaches the target depth, using a cuboidal box as the test object.

\begin{table}[h]
\centering
\caption{Comparison of grasp cycle time with and without ToF feedback for anticipatory gripper closure.}
\label{tab:reach2grasp}
\begin{tabular}{l|c|c}
\hline
\textbf{Condition} & \textbf{Mean Time (s)} & \textbf{Std. Dev. (s)} \\
\hline
Without ToF & 3.533 & 0.009 \\
With ToF & 3.271 & 0.015 \\
\hline
\end{tabular}
\end{table}

Table \ref{tab:reach2grasp} shows that ToF feedback reduces cycle time by approximately 0.26 seconds (7.4$\%$ reduction). By integrating ToF sensing into the approach phase, the gripper can initiate closure based on real-time proximity feedback rather than relying solely on position control. This anticipatory grasp strategy reduces cycle time by overlapping gripper closure with the approach motion, demonstrating that ToF-based depth perception provides sufficient accuracy and responsiveness for time-optimized grasping.

\subsection{Extraction}
For the second experiment, we demonstrate the value of ToFTouch by picking an item using only ToF to reach the proper depth. In cluttered environments, vision systems often provide an initial estimate of object location, but local depth corrections are needed as the EoAT approaches due to occlusions and sensing uncertainties. ToFTouch enables dynamic pre-contact depth adjustment during the pick sequence, reducing collision risks with neighboring objects.

In this experiment, we evaluate the system's ability to extract objects at varying depths relative to a nominal pick point. The object depth is varied using \SI{1}{\centi\meter} increments to simulate uncertainty in the initial depth estimate. When ToF feedback is enabled, we implement a PID controller in the z-axis using the z-axis ToF sensor. We use the same depth controller set point in all experiments. When ToF feedback is disabled, we command the robot to the nominal depth. The extraction setup is shown in Figure \ref{fig:robotiq}.

For each depth offset, we perform extraction ten times and record the success rate, defined as grasping the object without collision. The ToF sensors provide real-time depth corrections as the gripper approaches the target, compensating for the offset from the nominal pick depth. When ToF is enabled, we also record the difference between the nominal depth and the final z position of the gripper when the grasp occurs. This difference informs the accuracy of depth perception from the sensor. Table \ref{tab:extraction} presents the results, where ``Mean $\Delta$" represents the average measured deviation between the robot's actual displacement and the nominal depth, and ``SD" denotes the standard deviation of this measurement across trials.

\begin{table}[h]
\centering
\caption{Pick success and measured deviation across 1cm displacements from expected pick point with and without ToF feedback.}
\label{tab:extraction}
\begin{tabular}{c|c|c|c|c}
\hline
\textbf{Offset} & \textbf{No ToF} & \textbf{With ToF} & \textbf{Mean $\Delta$} & \textbf{SD} \\
\textbf{(cm)} & & & \textbf{(cm)} & \textbf{(cm)} \\
\hline
$-2$ & 0/10 & 9/10 & $-0.6$ & 0.4\\
$-1$ & 10/10 & 8/10 & 0.2 & 0.1\\
0 & 10/10 & 10/10 & 1.6 & 0.1\\
$+1$ & 10/10 & 10/10 & 2.5 & 0.07\\
$+2$ & 0/10 & 9/10 & 3.1 & 0.3\\
\hline
\end{tabular}
\end{table}

The results in Table \ref{tab:extraction} demonstrate that ToF feedback enables successful extraction across all tested depth offsets, including the extreme $\pm$\SI{2}{\centi\meter} conditions where the baseline approach without ToF fails due either to collision with the environment ($+$\SI{2}{\centi\meter}) or the object not being in the grasp ($-$\SI{2}{\centi\meter}). The primary failure mode when ToF is enabled is the sensor overheating which produces corrupt sensor readings.

We note the discrepancy between ``Mean $\Delta$" and ``Offset", which we would expect to match in a perfect world. We explain the discrepancy between the two by the lack of any calibration step and misalignment in ToF orientation when it is mounted on the gripper. More importantly, the trend across ``Mean $\Delta$" and small standard deviations suggest that ToF enables consistent depth correction performance even when the initial depth estimate differs by multiple centimeters from the true object location.

\subsection{Insertion}
In the final experiment, we explore the value of utilizing both ToF and tactile sensing modalities for insertion tasks in dense clutter. While vision systems struggle with occlusion when the gripper approaches tightly packed objects, the combination of pre-contact proximity sensing and contact force feedback enables continuous observations throughout the trajectory.

For this experiment, we set up a cluttered arrangement of boxes with minimal clearance to create a challenging insertion scenario for an oblong cardboard box. We use kinesthetic teaching to bring the robot arm through an insertion trajectory in which the arm inserts the oblong box into the available space between neighboring objects. We record ToF sensor data, tactile sensor data, and 3-axis force data from a 6 DoF ForceN force/torque sensor mounted between the arm and the gripper. The ToF sensors provide proximity feedback during approach, detailing information about the insertion location and clearance to surrounding items. The tactile patch detects contact forces during the insertion motion, providing information about extrinsic contacts with the environment. We also modify the ToF sensor angles to be 24$\degree$, 16$\degree$, and 8$\degree$ from vertical to provide better coverage of the environment for the insertion experiment.

\begin{figure}[htbp]
\centering
\includegraphics[width=\linewidth]{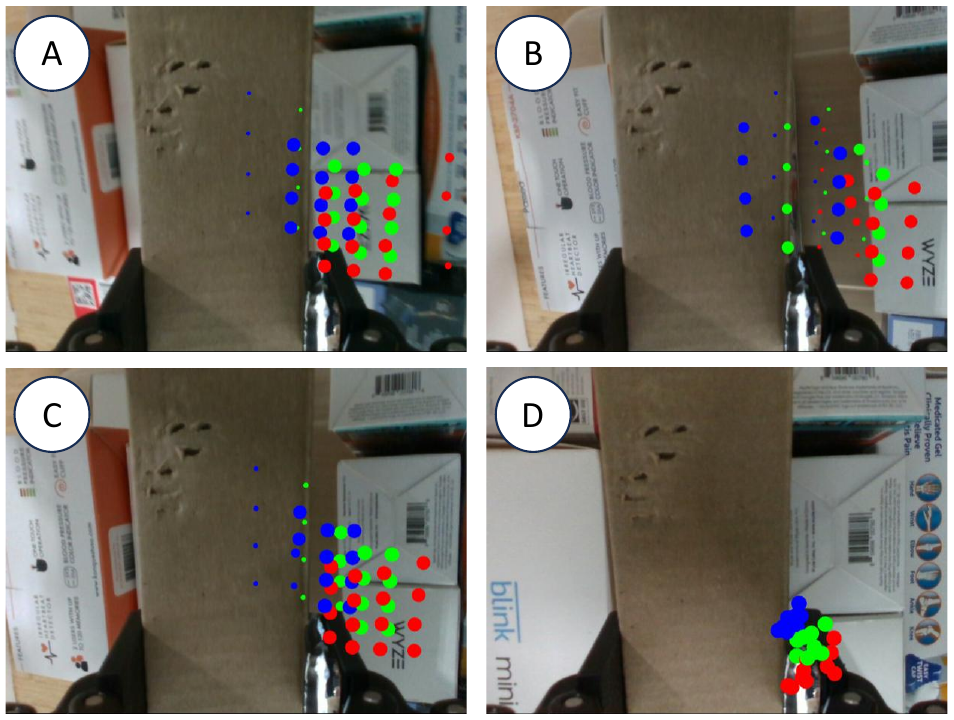}
\caption{ToF sensor data from three ToF sensors on ToFTouch during an insertion task into tight clutter. We use different ToF sensor angles for this task to provide better coverage of the environment. The angles are 24$\degree$, 16$\degree$, and 8$\degree$ from vertical for red, green, and blue points, respectively. Letters are in chronological order and correspond to Figure \ref{fig:insertion_plot}.}
\label{fig:insertion_images} 
\end{figure}

\begin{figure}[htbp]
\centering
\includegraphics[width=\linewidth]{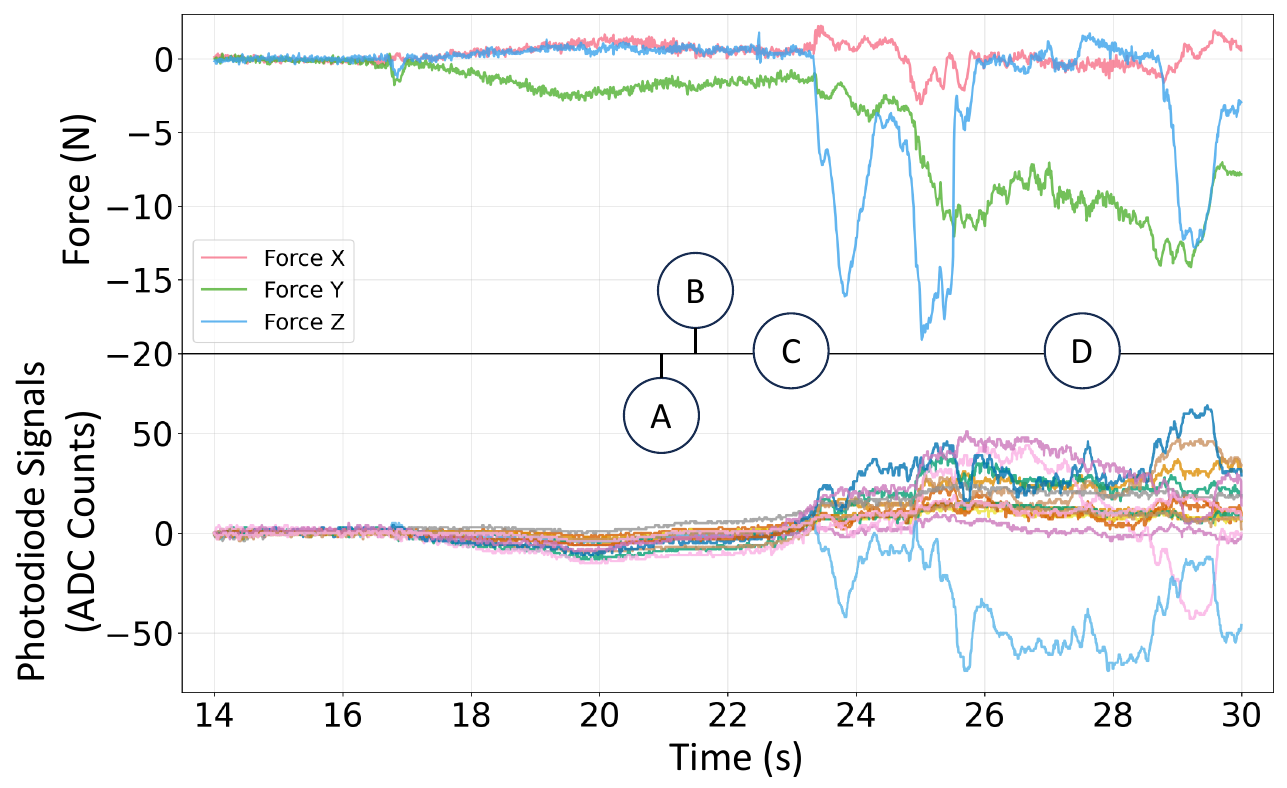}
\caption{Force measurements from a load cell (top), and tactile sensor readings (bottom) from ToFTouch during an insertion task into tight clutter. Each tactile signal corresponds to one LED and photodiode pair. Letter callouts correspond to Figure \ref{fig:insertion_images}.}
\label{fig:insertion_plot}
\end{figure}

Figures \ref{fig:insertion_images} and \ref{fig:insertion_plot} show the ToF sensor data and force measurements during the insertion task. The four images in Figure \ref{fig:insertion_images} visualize the ToF depth readings (red, green, and blue points for the three ToF sensors) at different stages of the insertion process. (A) shows detection of the top surfaces of neighboring items as the gripper approaches the scene. (B) shows detection of the empty cavity, with smaller circles indicating greater distance to the table surface compared to the larger circles representing closer proximity to the top surfaces of nearby items. (C) shows insertion into the cavity with ToF readings concentrated along the edge of the adjacent item. (D) shows all readings converging as the cardboard box is fully placed into the hole and the sensors become blocked by neighboring objects.

Figure \ref{fig:insertion_plot} compares 3-axis force measurements from the force/torque sensor to the raw analog readings from the tactile component on ToFTouch. A median filter of width 11 is applied to the tactile data. The force measurements remain near zero (with fluctuations due to changes in object orientation in space) until contact occurs, after which significant forces develop while nesting the cardboard box between neighboring items. The x-axis force reaches approximately \SI{-20}{\newton} and sustains y-axis forces of \SI{-10} to \SI{-15}{\newton}. The raw signals from the tactile patch show corresponding activation, with photodiode readings diverging from baseline values around the same time as force onset, indicating varying pressure patterns across the sensor array as the item makes contact with external objects. The combined ToF and tactile feedback enables the robot to monitor both pre-contact proximity and contact forces, demonstrating the value of multimodal sensing for dense packing tasks where visual occlusion limits traditional approaches.

\begin{figure}[htbp]
\centering
\includegraphics[width=\linewidth]{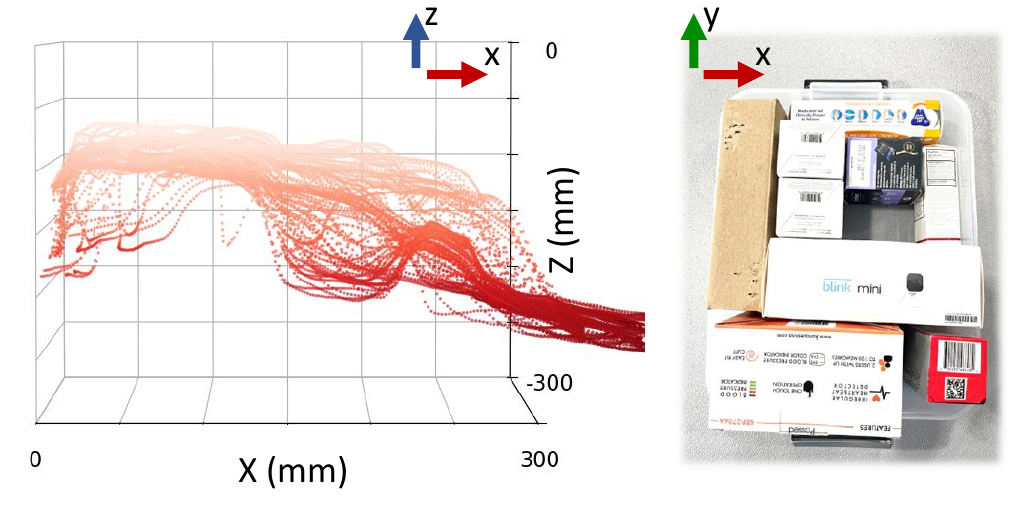}
\caption{Accumulated multizone ToF sensor data from ToF1 (as labeled in Figure \ref{fig:robotiq}) as a 3D point cloud (left) during a sweep over an insertion environment (right). The color gradient represents distance from the sensor. The data is shown at 60 Hz with a uniform filter of width 9 and samples less than \SI{50}{\milli\meter} filtered out.}
\label{fig:sweep}
\end{figure}

ToF data is richer in information when accumulated over many samples. To better understand the information accumulated over an insertion trajectory, we also sweep the Robotiq gripper while holding a rectangular object over a cluttered environment over multiple passes and plot the accumulated data from a single ToF sensor (ToF1, as labeled in Figure \ref{fig:robotiq}) in Figure \ref{fig:sweep}. We sample at 60 hz, filter out samples less than \SI{50}{\milli\meter}, and apply a uniform filter of width 9. The data show a clear depression where the hole in the environment is, while edges and small depth differences are noisy. Noisy edges appear because in a given zone, an edge may be detected by some photon emissions and not by others.

\section{Conclusion}
This paper presents ToFTouch, a multimodal sensor that integrates three multizone time-of-flight sensors with an optics-based tactile sensing array on a compact flexPCB design suitable for mounting on robotic grippers. The sensor provides both pre-contact proximity information and contact force feedback, addressing limitations of vision-only systems that suffer from occlusion during manipulation tasks and tactile-only approaches that lack spatial awareness before contact.

We characterize ToFTouch across three manipulation tasks that explore complementary roles for proximity and tactile sensing. In the pre-grasp adjustment experiment, ToF feedback enables anticipatory gripper closure during approach, reducing cycle time by 7.4$\%$ compared to position-based control. For extraction from clutter, ToF sensing enables successful object retrieval across depth offsets up to \SI{2}{\centi\meter} from the nominal pick point where vision-only approaches fail, with standard deviations below \SI{0.4}{\centi\meter} demonstrating consistent depth correction performance. In the dense packing insertion task, the combination of ToF proximity sensing and tactile force feedback enables rich observation before and during contact without visual information.

The results demonstrate that combining proximity and tactile sensing modalities can provide valuable information from a compact, fingertip view point that is not available to tactile-only or vision-only systems. We view this work as an initial characterization of fingertip-mounted multizone ToF and tactile sensing rather than a complete evaluation. Our quantitative results are based on a limited number of trials per condition, and a broader study with larger sample sizes, statistical analysis, and comparisons against additional proximity and tactile baselines would further strengthen these findings. Systematic calibration of the ToF sensors and characterization of their thermal behavior would also improve accuracy and reliability, addressing the sensor overheating and depth-offset discrepancies we observed. Future work could explore integration with learned control policies that leverage the rich spatial and force information or deployment on other end-effector embodiments, such as multi-fingered hands or palm surfaces, where the greater self-occlusion of dexterous hands makes local fingertip point-of-view proximity sensing particularly attractive.

\bibliographystyle{IEEEtran}
\bibliography{
    bib/references
    }

\addtolength{\textheight}{-12cm}   
\end{document}